\documentclass[conference]{IEEEtran}
\IEEEoverridecommandlockouts
\usepackage{cite}
\usepackage{amsmath,amssymb,amsfonts}
\usepackage{algorithmic}
\usepackage{graphicx}
\usepackage{textcomp}
\usepackage{multirow}
\usepackage{listings}
\usepackage{xcolor}
\usepackage{colortbl}
\usepackage[export]{adjustbox}
\usepackage[utf8]{inputenc}
\usepackage[english=american]{csquotes}
\usepackage{tabularx}
\usepackage{subcaption}
\usepackage{tcolorbox}
\usepackage{comment}
\usepackage{url}
\usepackage{balance}
\usepackage[ruled,vlined]{algorithm2e}
\usepackage{graphicx}
\usepackage{footnote}
\usepackage[perpage]{footmisc}

\usepackage[compact]{titlesec}         
\titlespacing{\section}{3pt}{3pt}{3pt} 
\AtBeginDocument{
  \setlength\abovedisplayskip{3pt}
  \setlength\belowdisplayskip{3pt}}
\newcommand{\Foutse}[1]{\textcolor{blue}{{\it [Foutse: #1]}}}

\def\BibTeX{{\rm B\kern-.05em{\sc i\kern-.025em b}\kern-.08em
    T\kern-.1667em\lower.7ex\hbox{E}\kern-.125emX}}
\begin{document}

\title{An Empirical Study of Challenges in Converting Deep Learning Models\\
}
\author{\IEEEauthorblockN{\Foutse{this is taking too much space, put all names in a single line separated by commas!} Moses Openja}
\textit{openja.moses@polymtl.ca}
\IEEEauthorblockA{\textit{Dept. of Computer Engineering} \\
\textit{Polytechnique Montréal, Canada}}
\and
\IEEEauthorblockN{Amin Nikanjam}
\textit{amin.nikanjam@polymtl.ca}
\IEEEauthorblockA{\textit{Dept. of Computer Engineering} \\
\textit{Polytechnique Montréal, Canada}}
\and
\IEEEauthorblockN{Ahmed Haj Yahmed}
\textit{ahmed.haj-yahmed@polymtl.ca}
\IEEEauthorblockA{\textit{Dept. of Computer Engineering} \\
\textit{Polytechnique Montréal, Canada}}
\and
\IEEEauthorblockN{Foutse Khomh}
\textit{foutse.khomh@polymtl.ca}
\IEEEauthorblockA{\textit{Dept. of Computer Engineering} \\
\textit{Polytechnique Montr\'{e}al, Canada}}
\and
\IEEEauthorblockN{Zhen Ming (Jack) Jiang}
\textit{zmjiang@cse.yorku.ca}
\IEEEauthorblockA{\textit{Dept. of Electrical Engineering \& \\ 
Computer Science, York University, Canada} 
}
}

\author{\IEEEauthorblockN{Moses Openja\textsuperscript{\dag}, Amin Nikanjam\textsuperscript{\dag}, Ahmed Haj Yahmed\textsuperscript{\dag}, Foutse Khomh\textsuperscript{\dag}, and Zhen Ming (Jack) Jiang\textsuperscript{\ddag}}
\IEEEauthorblockA{
\{openja.moses,amin.nikanjam,ahmed.haj-yahmed,foutse.khomh\}@polymtl.ca
, zmjiang@cse.yorku.ca}

\IEEEauthorblockA{\textsuperscript{\dag}Polytechnique Montréal, Montreal, Quebec, Canada \\
\textsuperscript{\ddag}York University, Toronto, Ontario, Canada 
}
}
\maketitle

\begin{abstract}
There is an increase in deploying Deep Learning (DL)-based software systems in real-world applications. Usually DL models are developed and trained using DL frameworks like TensorFlow and PyTorch. Each framework has its own internal mechanisms/formats to represent and train DL models (deep neural networks), and usually those formats cannot be recognized by other frameworks. Moreover, trained models are usually deployed in environments different from where they were developed. To solve the interoperability issue and make DL models compatible with different frameworks/environments, some exchange formats are introduced for DL models, like ONNX and CoreML. However, ONNX and CoreML were never empirically evaluated by the community to reveal their prediction accuracy, performance, and robustness after conversion. Poor accuracy or non-robust behavior of converted models may lead to poor quality of deployed DL-based software systems. We conduct, in this paper, the first empirical study to assess ONNX and CoreML for converting trained DL models. In our systematic approach, two popular DL frameworks, Keras and PyTorch, are used to train five widely used DL models on three popular datasets. The trained models are then converted to ONNX and CoreML and transferred to two runtime environments designated for such formats, to be evaluated. We investigate the prediction accuracy before and after conversion. Our results unveil that the prediction accuracy of converted models are at the same level of originals. The performance (time cost and memory consumption) of converted models are studied as well. The size of models are reduced after conversion, which can result in optimized DL-based software deployment. We also study the adversarial robustness of converted models to make sure about robustness of deployed DL-based software. Leveraging the state-of-the-art adversarial attack approaches, converted models are generally assessed robust at the same level of originals. However, obtained results show that CoreML models are more vulnerable to adversarial attacks compared to ONNX. The general message of our findings is that DL developers should be cautious on deployment of converted models that may 1) perform poorly while switching from one framework to another, 2) have challenges in robust deployment, or 3) run slowly, leading to poor quality of deployed DL-based software, including DL-based software maintenance tasks, like bug prediction.
\end{abstract}




\section{Introduction} \label{sec:intro}
Nowadays, we are observing an increasing deployment of Deep Learning (DL)-based software systems in real-world applications, from personal banking to autonomous driving\cite{heaton2020applications,Moses2022:Docker}. Different easy-to-use Python-based frameworks are developed to help practitioners write their own DL codes, like TensorFlow/Keras~\cite{keras,abadi2016tensorflow} and PyTorch~\cite{torch}. A (or multiple) trained DL model(s) must be deployed in a DL-based software system to provide the necessary prediction service. There are several stages in development of DL-based software prior to deploying a trained model: from data collection, labeling and designing Deep Neural Networks (DNNs) to training/testing the model. The development team encodes the network structure of the desirable DL model (a DNN in particular) and the process by which the model learns from a training dataset. Hyperparameters and runtime settings (e.g., random seed) should also be configured prior to training the model on the selected dataset. Afterward, a validation/testing stage is performed for evaluating the prediction accuracy of the trained models. There might be some back-and-forth steps as well to improve the quality of the model. Finally, the model becomes ready to be deployed for application, for example on web or mobile platforms. To do so, the trained model is converted to some format and stored as one or multiple files. 

A DL framework may support multiple options to store (or export) models, e.g. TensorFlow provides \textit{checkpoints} during the training and saving trained model using \textit{SavedModel} and \textit{HDF5} \cite{TF-save}. However, each framework has its own representation for a model with customized primitives, which are not necessarily compatible among different frameworks. Since the deployment environment is likely different from the development, it is expected that the trained model in different frameworks can be easily ported and reused by another framework or in a different environment \cite{liu2020enhancing,InteroperableAI}. This interoperability requirement enforces that the prediction accuracy, performance, and quality of the ported (converted) model be the same as (or reasonably similar to) the original trained model. Deploying a DL model with unexpected poor accuracy or quality in a software system may affect the quality of final decisions having severe consequences, especially in the context of safety-critical systems. It is therefore important to raise the awareness of development teams about the importance of interoperability requirements of trained DL models during deployment.

To address such interoperability requirements, different conversion formats and conversion techniques are introduced for DL formats, e.g., Open Neural Network Exchange (ONNX) \cite{ONNX:2021} and CoreML \cite{CoreML:2022}. The converted DL model should be easily transferred into other inference/development environments due to deployments requirements, infrastructure limitations, or security requirements (data privacy). On the other hand, these conversion formats are necessary since development of multiple DL frameworks is inevitable: they provide their users with diverse DL development experience including model training performance, DevOps productivity and release engineering~\cite{openja2020:release}, and debugging. According to a recent report \cite{kaggle-2021}, TensorFlow/Keras were selected by about half of the developers in 2021 while PyTorch was employed by about 34\% of users.  


Although these formats are currently used by researchers/practitioners for converting DL models from different frameworks, to the best of our knowledge there is no study that empirically evaluated the prediction accuracy, inference time, memory usage, and robustness of those formats. An empirical study in this direction can identify the challenges of the current conversion techniques. Also future research can use such study to drive the various DL software maintenance tasks, such as practical fault detection and verification methods, to support debugging the conversion techniques and correctness of the deployed DL software. Therefore, we define our Research Questions (RQs) as follows:\\
\begin{itemize}
    \item \textbf{Prediction accuracy after conversion.} \\ \textbf{RQ1:} Given the same runtime configuration, what are the differences of prediction accuracy between an original trained model and its converted one? 
    \item \textbf{Performance of converted models\footnote{By performance, here, we mean cost-related metrics such as time and CPU resources.}.}\\ \textbf{RQ2:} 
    How does a DL model perform after conversion? Is there any difference in the inference time and size of the converted model? 
    \item \textbf{Adversarial robustness of converted models.} \\ \textbf{RQ3:} Do converted models from a DL framework exhibit the same adversarial robustness as the original ones after conversion?
\end{itemize}

In this paper, we conduct an empirical study on the state-of-the-art conversion format, i.e., ONNX and CoreML, using different DL models trained in two widely used frameworks of Keras (an API on the top of TensorFlow) and PyTorch. We aim to investigate the performance and robustness of converting trained DL models, its impact on DL software development, maintenance/deployment processes, and  providing practical insight for both SE and AI communities on the different aspects that should be improved. To summarize, this paper makes the following contributions: 
\begin{itemize}
	\item We present the first experimental study on effectiveness and performance of ONNX and CoreML conversion formats for trained DL models. 
    \item We report the difference in predictions between ONNX and CoreML models: the error for ONNX is quite minimal compared to CoreML while in the case of CoreML, the error differs dramatically across Keras and PyTorch.
    \item We observe variations in inference time of the converted models: ONNX models are faster than CoreML models. 
	\item The converted models are generally robust against adversarial attacks at the similar level of original models. However, our assessment reveals that conversion to CoreML is more vulnerable to adversarial threats than ONNX conversions.
	\item We release a replication package including our detailed results \cite{ReplicationPackage}, that can be used as a benchmark for other studies on exchanging trained DL models. 
\end{itemize}

The rest of this paper is organized as follows. While Section \ref{background} briefly reviews ML conversion formats, Section \ref{methodology} details the methodology followed in our study. We report the experimental results for each of our RQs in Section \ref{results}. Section \ref{discussion} presents a discussion on our finding and highlights future research opportunities. We conclude the paper in Section \ref{conclusion}.

\begin{figure*}
\centering
\includegraphics[width=0.83\textwidth]{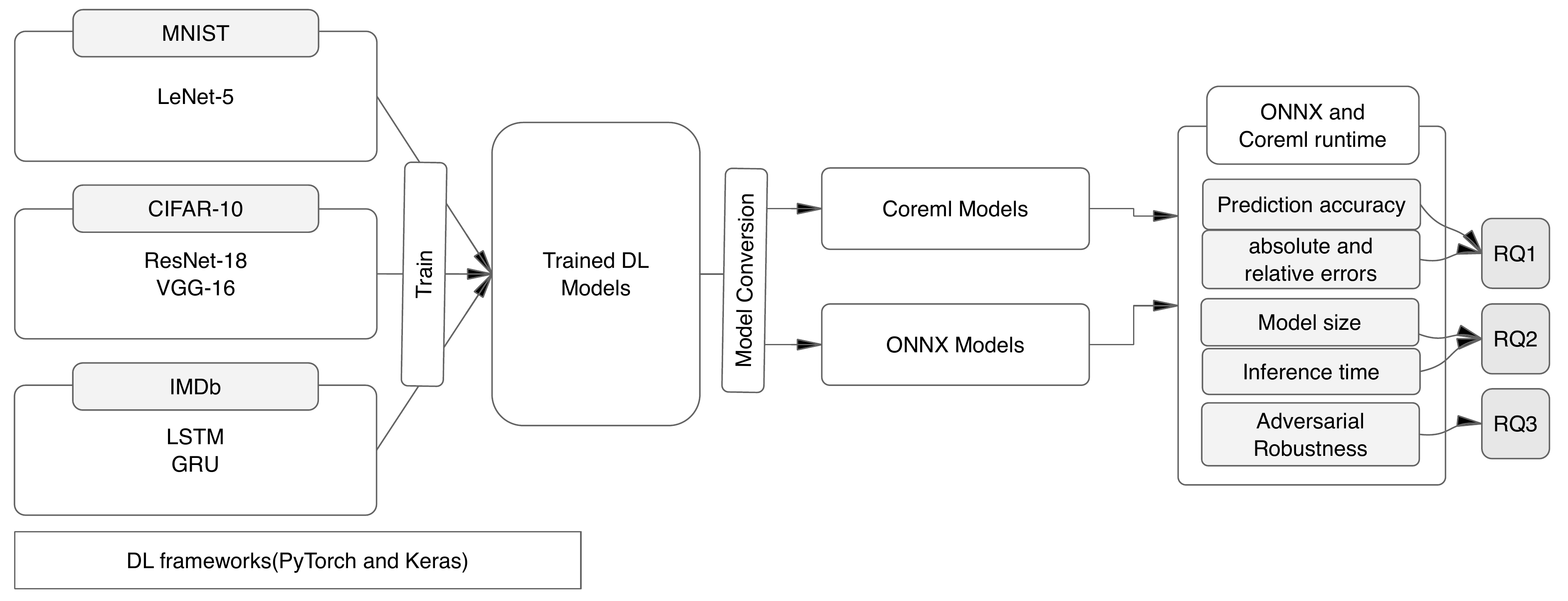}
\caption{Our methodology to train, convert, and evaluate converted DL models.}
\label{fig:overview}
\end{figure*}





\section{Background}\label{background}
In this section, we briefly review conversion formats for Machine Learning (ML) models.

ONNX is an open source machine-independent format for representing ML models \cite{ONNX:2021}. The ONNX model is an expandable computational graph equipped with operators, and standard data types offering a uniform representation for different frameworks. It is widely employed for converting DL models and has been actively maintained by and contributed from open source communities \cite{jin2020compiling}. Moreover, its official tutorial for converting DL models to ONNX format is available online \cite{ONNX:Github2021,ONNX:Tutorial2021}. 

CoreML is an Apple framework to integrate ML models into applications by providing a unified representation for all types of models \cite{CoreML:2022,CoreMLRepo:2022}. It optimizes the on-device performance of models by leveraging the CPU, GPU, and Apple Neural Engine (ANE) while minimizing the memory footprint and power consumption. Besides ONNX and CoreML, other conversion formats were introduced as well. 

Predictive Model Markup Language (PMML) is a java based library for moving Apache Spark, R and Scikit-Learn models from “lab” to “factory” \cite{PMML:2021} with a Python library for converting Scikit-Learn pipelines \cite{pythonPMML:2021}. H2O is an open-source java-based scalable ML and predictive analytics platform for building ML models on big data with easy productionalization \cite{H2O:2021}. It allows for converting the built models to Plain Old Java Object (POJO) or a Model ObJect, Optimized (MOJO) \cite{productH2O}.


\section{Study Design} \label{methodology}

In this section, we introduce the methodology of our study to assess converted DL models. Figure \ref{fig:overview} illustrates an overview of our study in this paper. There are three main steps to answer our RQs: 1) training models, 2) converting them to ONNX and CoreML formats, and 3) evaluating the converted models. First, we train different DL models and investigate their prediction accuracy across different frameworks. To do so, two widely-used and well-known frameworks, i.e., PyTorch \cite{torch} and Keras \cite{keras}, are selected. Regarding DL models, similar to other studies \cite{guo2019empirical,pham2020problems}, we choose five popular models: LeNet-5 \cite{lecun1998gradient}, ResNet-18 \cite{he2016deep}, and VGG-16 \cite{simonyan2014very}, LSTM \cite{lstm} and GRU \cite{gru}. For training and then evaluating these models, we have employed three famous publicly available datasets: IMDb~\cite{IMDb:2021}, CIFAR-10~\cite{krizhevsky2009:learning}, and MNIST~\cite{lecun1998:mnist}.

After training each of the five selected DL models, we converted the models from the respective frameworks (PyTorch and Keras) to ONNX and CoreML formats. Then, we employ ONNX Runtime (ORT)~\cite{ONNX:runtime} and CoreML Tools Python package (a.k.a \texttt{coremltools}) \cite{CoreMLRepo:2022} to run and perform experiments with converted models. Each experiment is repeated 10 times and the average values are recorded. It was acknowledged in the literature that multiple DL training runs with identical configuration may result in different results. So, multiple runs are necessary to analyze the variance and increase the reliability of results~\cite{pham2020problems}.  

\subsection{DL Frameworks}
There exist a number of DL frameworks for developers and researchers to design, train, test, and deploy a variety of deep models across the industry and academia. For this study, we selected two state-of-the-art frameworks, i.e., PyTorch 1.9.0 from Facebook and Keras 2.6.0 (running on the top of TensorFlow 2.6.2) from Google. Each of these frameworks defines its own network structure and format for creating, training, and saving models. For instance, PyTorch follows a dynamic computational paradigm, whereas TensorFlow/Keras follows the static computational graph paradigm. It should be noted that Keras, as a set of DL APIs developed in Python, was originally designed to support multiple backends, including TensorFlow, Theano, and PlaidML, however as of version 2.4, only TensorFlow is supported \cite{Keras-release}.

\subsection{Models and Datasets}\label{subsec:model-dataset}
We choose five popular models widely used in the DL community. Among Convolutional Neural Networks (CNNs), we selected LeNet-5, ResNet-18, and VGG-16. These models are mainly designed for image classification. To keep the set of models diverse enough, LSTM and GRU are chosen from deep Recurrent Neural Networks (RNNs). The last two are widely used in Natural Language Processing (NLP) (like sentiment analysis) and sequence prediction (like time series).
 
For training and then evaluating the models, we have employed three famous publicly available datasets: i.e., IMDb~\cite{IMDb:2021} for LSTM and GRU, CIFAR-10~\cite{krizhevsky2009:learning} for ResNet-18 and VGG-16, and MNIST~\cite{lecun1998:mnist} for LeNet-5 respectively. MNIST is a set of gray-scale images used for recognizing handwritten digits. CIFAR-10 is collected from colored images for object classification, like airplane, automobile, and bird. IMDb is a set of text-based movie reviews from the online database IMDb \cite{imdb-dataset}, which is widely employed for text sentiment classification in NLP tasks. The selected datasets are very popular in the DL community. 

\subsection{Model Training and Conversion}
In this step, for each of the selected DL models and the datasets described in Subsection~\ref{subsec:model-dataset}, we first train and test DL models in the corresponding frameworks. We train our DL models on the datasets using Keras and PyTorch. Multiple combinations of hyperparameters for each model are tested in the training step to obtain a proper training accuracy on each framework preventing overfitting/underfitting issues. The selected models are trained using the identical hyperparameter setting such as the optimizer, the learning rate, the batch size, for both DL frameworks (PyTorch or Keras). Each experiment was repeated 10 times and the average values were recorded.
 
After training each of the five DL models, we convert the models from the respective frameworks to ONNX and CoreML formats. In Keras, we have used \texttt{tf2onnx} 1.9.1, an official toolset for converting TensorFlow and Keras models to ONNX \cite{tf2onnx}. To convert ONNX models in PyTorch, \texttt{torch.onnx} is used. \texttt{torch.onnx} consists of a set of PyTorch built-in API for working with ONNX format~\cite{torch.onnx}. To evaluate the converted ONNX models, we employ ONNX Runtime (ORT) 1.10.0 \cite{ONNX:runtime}. ORT is an engine, developed in C++, used to deploy ONNX models into production with high performance. It is capable of integrating with hardware-specific libraries using a flexible interface, so it performs inferences efficiently across various platforms and hardware, e.g., Windows, Linux, and Mac, on both CPUs and GPUs. ORT supports DL as well as other ML models. To convert trained models to CoreML and run them, we used  \texttt{coremltools} 5.0b3 \cite{CoreMLRepo:2022}, which is a Python package acting as the primary tool to convert third-party models to CoreML. \texttt{coremltools} supports converting trained models from various libraries and frameworks to the CoreML format and making predictions using converted models.






\subsection{Evaluation Metrics}
This section describes the metrics used to evaluate the converted models. The prediction accuracy, inference time, model size, and adversarial robustness of converted models are assessed and compared to the original models.
 
\textbf{Accuracy in Training and Conversion.} To train models, we first ensure the same runtime configuration across different frameworks. We attempt to train the models with different combinations of hyperparameters on each framework, and monitor the training and validation accuracy. At the end, one combination is selected as reported in the paper, which achieves acceptable training accuracy for all selected frameworks. Then, we assess the prediction accuracy on the testing data for each model/dataset. After converting models to ONNX and CoreML, we evaluate the prediction accuracy of converted models on the testing data to identify any difference that may be caused by the conversion across different frameworks. Poor accuracy or wrong decisions of the deployed model may affect the functionality of the system.

To further assess the accuracy of converted models, we calculate the difference between the predictions generated by the converted model (ONNX and CoreML models) and their respective original models (in PyTorch and Keras). To evaluate the predictions of models, first, we compare the output of the last layer per pair of neural network models, i.e., original vs. converted models. We report the difference between the last layers' outputs (i.e., predictions) of original and converted models as the error. Two types of errors are reported: absolute and relative error. The absolute error measures the amount of error in the predictions of the converted model as the difference between the predictions of the converted model and the original model (before the conversion). Formally, it is defined as: 
\begin{equation}
    \textit{Absolute error} = \Delta y = y_c - y_o
\end{equation}
where $y_c$ is the prediction results (can be a vector) of the converted model (i.e., ONNX or CoreML), and $y_o$ is the prediction results of the original model (i.e., in Keras or PyTorch). When $y$ is a vector, we report the average error over all elements of the vector. Similarly, the relative error is defined as the ratio of the absolute error of the converted model to the predictions of the original model:
\begin{equation}    
    \textit{Relative error} = \frac{\Delta y}{y_o} 
\end{equation}

\textbf{Performance in Training and Conversion.} Intuitively, converting a model to another format results in some change in the structure of the model. Therefore, the performance of the model, i.e., running time or memory usage, might be affected. After training, we evaluate the performance of trained models during prediction tasks on the testing data. To do so, we assess the size of the model (memory used to store the model), and inference time (and load time for converted models). Then, we repeat our evaluations on the converted models, and compare the results to those of the original. However, to ensure that this has a limited impact on the quality of the deployed software system, this difference should not be significant. For example, poor inference time of a deployed model may result in longer response times and affect the quality of DL-based software.
 
\textbf{Adversarial Robustness in Conversion.} Resilience and security of a trained model is determined by robustness evaluation. Deploying non-robust may lead to degrading the security of DL-based software systems. Among different robustness properties, we evaluate the adversarial robustness of models in this paper. The adversarial robustness can be simply described as follows: if a model M misclassifies an input $x'$, which is close to a given input $x$, then $x'$ is considered as an adversarial example of $x$ and M is considered not adversarial robust. To speak formally, the adversarial robustness of a classifier can be evaluated by \textit{d-local-robustness} at an input $x$ using the following criterion:
\begin{equation}
    \forall x: dist(x,x') \leq d \implies \mathcal{C}(x) = \mathcal{C}(x'),
\end{equation}
where $dist$ indicates the distance between two input samples, and $\mathcal{C}$ is the indicated class for $x$ by the classifier. In this paper, we follow the state-of-the-art techniques in DL to generate adversarial examples using adversarial attacks on DL models \cite{goodfellow2014explaining, brendel2017decision} to assess the robustness of converted models and compare them to originals. Moreover, in this paper, the adversarial robustness assessment is performed not only for CNN models but also for RNNs by following state-of-the-art techniques \cite{alzantot-2018}. 


\begin{figure*}[t]
     \centering
     \begin{subfigure}[b]{0.45\textwidth}
         \centering
         \includegraphics[width=\textwidth]{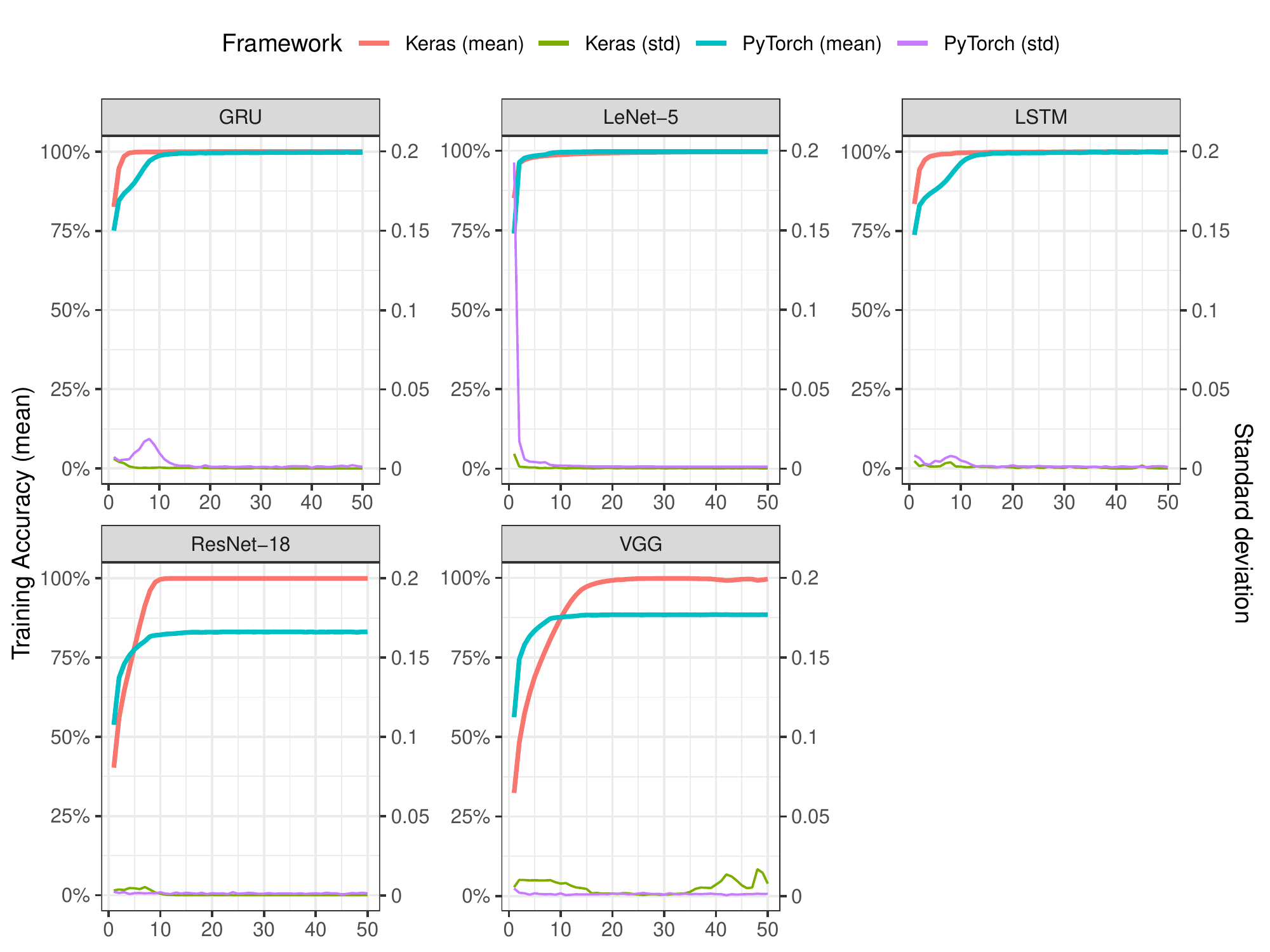}
         \caption{Training Performance}
         \label{fig:Training}
     \end{subfigure}
     \hfill
     \begin{subfigure}[b]{0.45\textwidth}
         \centering
         \includegraphics[width=\textwidth]{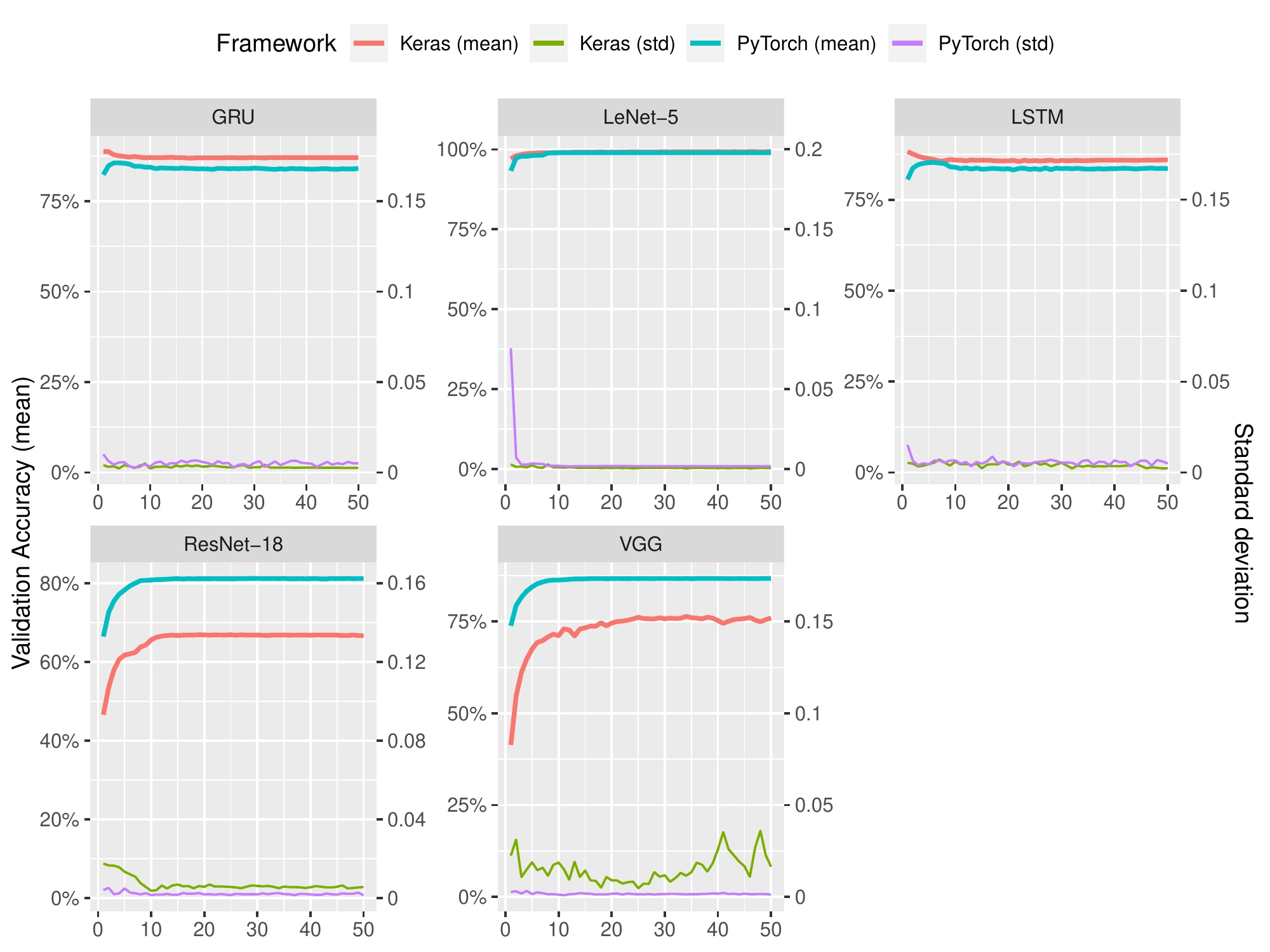}
         \caption{Validation Performance}
         \label{fig:Validation}
     \end{subfigure}
        \caption{The training and validation performance of the selected model with respective frameworks. 
        }
        \label{fig:validation}
\end{figure*}

\section{Empirical Results} \label{results}
In this section, we first briefly introduce the experimental environment, then we detail the experiments and results obtained to answer our RQs.





\textbf{Experimental Environment.}
To train models, we use a high performance computer running CentOS-7 on a 1.70 GHz Intel Xeon Bronze 3104 CPU with 64 GB main memory equipped with a NVIDIA GeForce RTX 2080 Ti GPU. All other experiments (i.e., converting trained models and assessing for both ONNX and Coreml) are run on a Macbook Air laptop with macOS 12.2.1 on a 1.6 GHz Dual-Core Intel Core i5 CPU with 16GB main memory. The choice for using a Mac machine is due to the limitation of coreML models which can be run only on Mac systems \cite{pypiCoreML:2022}. 


\subsection{RQ1: Prediction accuracy of converted models}
\subsubsection{Training accuracy}
We have trained five DL models on three datasets in Keras and PyTorch: LSTM and GRU on IMDb, Restnet-18 and VGG-16 on CIFAR-10, and LeNet-5 on MNIST. 
The runtime configuration per model is the same for different frameworks. For example, we have used identical learning rate (i.e., 0.05), training epochs (i.e., 50), optimizer (i.e., SGD), batch size (i.e., 128) for LeNet-5 on both frameworks. Each model is trained 10 times in each framework and the average results is reported for comparison. We have reported the hyperparameters used for training each DL model in our replication package.

Figure~\ref{fig:validation} visualize training and validation plots of all five models given identical configurations on different DL frameworks. Overall we can observe that there is a closely similar training behavior shown by both frameworks except for Keras that demonstrated much higher training accuracy compared to PyTorch in our study for VGG-16 and ResNet-18. 

For each DL model, we evaluate the prediction on the testing dataset after the training. We repeat predictions on all 10 instances for each model and record the average accuracy. The result is presented in Table \ref{tab:projects_category}. The prediction accuracy of LeNet-5, VGG-16, and LSTM is similar across Keras and PyTorch but this is not the case for ResNet-18 and GRU. One can justify the results since DL frameworks employ different computing libraries having various implementations of operators (like calculating gradients, operations of layers, or optimizations). Besides, the non-deterministic nature of the training process eventually makes the weights/biases on the same layer different from each other while the structure is the same. Moreover, similar observations (and results) were reported in the literature \cite{guo2019empirical}. 

\begin{table}[t!]
    \large
    \caption{The average prediction accuracy of original and converted models over the testing datasets per framework.}
    
    \label{tab:projects_category}

   \begin{adjustbox}{width=0.8\columnwidth,center}
    \begin{tabular}{l c | c|c|c}

    \multirow{3}{*}{\textbf{Model}}& \multirow{3}{*}{\textbf{Framework}}&\multicolumn{3}{c}{\textbf{Accuracy (\%)}}\\
    & & \textbf{Original} & \textbf{ONNX} & \textbf{CoreML}\\
    
    \hline
     \multirow{2}{*}{\textbf{LeNet-5}}&\textbf{Keras}& 99.85 & 99.85 & 99.85\\
    
    &\textbf{PyTorch}& 98.90 & 98.90 & 98.90\\
    \hline
    
     \multirow{2}{*}{\textbf{ResNet-18}}&\textbf{Keras}& 85.99 & 85.99 & 86.01\\
    
    &\textbf{PyTorch}& 81.69 & 81.69 & 81.72\\
    \hline
    
     \multirow{2}{*}{\textbf{VGG}}&\textbf{Keras}& 88.66 & 88.66 & 88.65\\
    
    &\textbf{PyTorch}& 81.40& 81.40 & 81.40\\
    \hline
    
     \multirow{2}{*}{\textbf{LSTM}}&\textbf{Keras}& 81.32 & 81.32 & 81.32\\
    
    &\textbf{PyTorch}& 74.22 & 74.22 & 74.22\\
    \hline
    
     \multirow{2}{*}{\textbf{GRU}}&\textbf{Keras}& 84.98 & 84.98 & 84.98\\
    
    &\textbf{PyTorch}& 71.09 & 71.09 &71.09\\
    
    \end{tabular}
    \end{adjustbox}
\end{table}


\subsubsection{Accuracy of converted models} \label{accuracyConverted}
We have converted trained models into ONNX and CoreML from the framework in which they were trained. For each trained model per framework, we convert all 10 instances of the model to ONNX and CoreML. So, we have 10 converted instances for each model. All the reported results are averaged over 10 instances per model in each framework. 

Table \ref{tab:projects_category} reports the prediction accuracy of converted models over testing dataset. The accuracy of original models are also reported in this table for comparison. As it is expected, converted models demonstrate almost similar accuracy compared to the original ones. While the differences are negligible (or almost zero) for ONNX models, the largest offset obtained in our experiments belong to CoreML format. This may be caused by changing the structure of the model during conversion that leads to missing some information during the conversion or potential numerical defects in \texttt{coremltools} as the runtime environment. Further investigation is essential to identify the root cause of such an offset.

Table \ref{tab:accuracy-summary} reports absolute and relative errors for five converted models from Keras and PyTorch. Overall, the prediction errors for the DL models in ORT (running ONNX models) is quite minimal compared to when the similar model is exported to \texttt{coremltools} as observed by their both absolute and relative errors. The prediction errors for the models converted to CoreML differ dramatically across the DL frameworks and the type of models: there is a big difference between Keras and PyTorch in absolute errors of VGG-16 and relative errors of ResNet-18. Moreover, RNN models tend to perform much better (with lower error) in CoreML compared to CNN models. A possible reason can be due to incorrect scaling factors during the sampling (upsampling or downsampling) operations during the conversion, which is a common operation specifically in CNNs for performing image segmentation. 

We also, in Table \ref{tab:accuracy-summary}, report \textit{misclassification} observed between original and converted models. To measure misclassification rate, for each sample in the model’s testing dataset, we check whether the sample is labeled the same by the original and converted models or not. The percentage of misclassified samples are reported accordingly. While this rate is zero for most of the converted models, for ResNet-18 and VGG-16, CoreML models show high misclassification rates compared to ONNX. This ratio indicates the extent to which prediction errors of the converted models can affect their functionality of labeling input samples as a classifier. It is possible that negligible prediction errors result in different labels for the same input. This means that the deployed DL-based software system behaves differently from the trained/tested system on the same input. Different decisions of a system for the same input, can affect its functionality as well by introducing nondeterministic behavior. 

\begin{table*}[t]
    \centering
    \large
    \caption{The comparison of mean and standard deviation (in brackets) of absolute and relative errors of output the converted DL models to ONNX and CoreML per framework. 
    }
    \label{tab:accuracy-summary}
   \begin{adjustbox}{width=0.9\linewidth}
    \begin{tabular}{l c | c| c| c| c| c| c}
    
    \multirow{2}{*}{\textbf{Model}}& \multirow{2}{*}{\textbf{Framework}}&\multicolumn{2}{c|}{\textbf{Absolute Error}}&\multicolumn{2}{c|}{\textbf{Relative Error}}&\multicolumn{2}{c}{\textbf{Misclassification}}\\
    & &\textbf{ONNX}&\textbf{CoreML}  &\textbf{ONNX}&\textbf{CoreML}&\textbf{ONNX(\%)}&\textbf{CoreML(\%)}\\ 
    
    \hline
     \multirow{2}{*}{\textbf{LeNet-5}}&\textbf{Keras}&6.7E-09(8.8E-09)	&1.4E-02(2.9E-02) & 9.5E-08(4.5E-07) & 1.4E-01(3.1E-01) & 0(0) & 0(0)\\
    
    &\textbf{PyTorch}&8.9E-07(1.1E-06) &1.0E-02(1.3E-02)&1.1E-06(5.9E-05)&	1.3E-02(4.5E-01)& 0(0) & 0(0)\\
    
    \hline
    
     \multirow{2}{*}{\textbf{ResNet-18}}&\textbf{Keras} & 1.5E-07(5.3E-07) &1.6E-01(2.0E-01) & 5.1E-06(4.4E-06) & 1.6(2.0)	& 0(0) & 0.70(0.41)\\
    
    &\textbf{PyTorch}&2.0E-06(1.7E-06)	&6.3E-03(5.7E-03)&4.9E-06(1.6E-04)&	0.02(1.1) &0(0) &0.12(0.11) \\\hline

     \multirow{2}{*}{\textbf{VGG}}&\textbf{Keras}&1.4E-07(3.4E-07)	& 1.3E-01(2.5E-01)	& 2.1E-06(1.9E-06) & 5.2E+05(4.1E+06) & 0(0)& 0(0) \\ 
     
     &\textbf{PyTorch}&4.2E-07(8.0E-08)	&6.9E-05(5.1E-06)&4.2E-06(5.1E-06)&1.7E-03(9.4E-03) & 0(0) & 0.37(0.25)\\
    
    \hline
    
     \multirow{2}{*}{\textbf{LSTM}}&\textbf{Keras}&4.3E-08(7.8E-08)	& 6.3E-08(2.4E-07) & negligible & 7.5E-06(3.2E-05)& 0(0)& 0(0) \\
    
    &\textbf{PyTorch}&5.7E-08(8.4E-08) & 3.1E-08(7.3E-08)&6.8E-07(2.5E-06)&	1.4E-07(2.9E-07)& 0(0) & 0(0) \\
    
    \hline
    
     \multirow{2}{*}{\textbf{GRU}}&\textbf{Keras}&4.5E-08(1.6E-07)	& 5.1E-08(2.4E-07)	& negligible & 5.1E-06(2.5E-05)& 0(0) & 0(0)\\
    
    &\textbf{PyTorch}&5.5E-08(5.4E-08)	& 3.2E-08(4.6E-08)&4.2E-07(1.1E-06)&	1.4E-07(1.9E-07)&0(0) &0(0) \\
    
    \end{tabular}
    \end{adjustbox}

\end{table*} 

\begin{tcolorbox}[colback=blue!5,colframe=blue!40!black]
\textbf{Findings:} While the accuracy difference between the original and converted models are evaluated to be negligible, ONNX models look more accurate than CoreML according to our results.

\textbf{Challenges:} Further analysis is necessary to investigate the effect of model's structure and runtime environment in the prediction performance of converted models. Moreover, different decisions of the converted models for the same inputs (quantified by \textit{misclassification}) can be very harmful for the functionally of DL-based software systems.
\end{tcolorbox}

\subsection{RQ2: Performance of Converted Models}
Intuitively, converting the trained model to ONNX or CoreML formats may result in a different structure of the model (i.e., layers are not necessarily implemented in exactly the same way across frameworks) and therefore the model size is likely to be modified. Since the model structure is changed, the inference process can be changed leading to different inference times. In this way, for example, long response time can be observed in the deployed DL-based software affecting its quality while it is not observed during testing trained models. 

To answer this RQ, we evaluate the inference time and size (used memory to store the model) of converted and original models and compare them. Normally, the total inference time is the sum of time consumed for model loading and inference (i.e., load model + perform inference). However, in practice the model is loaded only once. We reported both the inference and load time for all of our models. As stated earlier, all the experiments were repeated 10 times per model and the average is reported.

Table \ref{tab:inference-time-summary} reports the inference time of original and converted models. Regarding the time costs of inference, the difference among frameworks is significant. It takes much more time on PyTorch to predict the outcomes of RNN models (both LSTM and GRU). For example, in the case of LSTM, the prediction lasted about 3.79 seconds on PyTorch while it takes only 0.17 seconds on Keras. This is mainly because PyTorch dynamically loads the data along with the graph building at each batch, without feeding them in advance. In this way, PyTorch inevitably generates a large number of temporary variables in an instant, leading to long inference time. For VGG-16, the inference time is almost the same for both frameworks, we have the same observation for prediction accuracy. Two other CNNs, i.e., LeNet-5 and ResNet-18, demonstrate a different behavior where Keras performs the inference much faster than PyTorch.

For converted CNNs, the inference time of the DL models in the ORT environment is faster than that of the original model while for CoreML it is slower. Contrary to the PyTorch models, Keras models are not uniform when converted to ONNX and CoreML. For example, for the LeNet-5, the inference time is much slower in CoreML and much faster in ONNX when the employed framework is Keras. Similar to the CNN models, RNN models perform much faster (inference time) in ORT than \texttt{coremltools} and the original frameworks. However, the inference time of the DL models in CoreML format is not uniform for different DL frameworks. This phenomenon leads to poor quality DL-based software systems with longer response times. Longer response times of software systems may cause lower user satisfaction and poor productivity among users, that may lead the user to discontinue using the software system.
 
The size of original and converted models are reported in Table \ref{tab:memory-summary}. This is the amount of memory used to store the model in MB. The size of CNN models reduces when converted from DL frameworks to ONNX format. Overall, the model sizes in CoreML are slightly less compared to the ONNX ones. By comparing the model sizes across the DL frameworks, one can conclude that Keras models are much bigger in size compared to the PyTorch models. Similar to the CNN models, the size of the RNN models reduces when converted from DL frameworks to ONNX and CoreML format. However, contrary to the above conclusion, the model size in CoreML is slightly higher than ONNX. Also, by comparing across the DL frameworks, RNN models trained in Keras are much bigger in size compared to peer models in PyTorch.

\begin{table*}[t]
    \centering
    \caption{The comparison of mean and standard deviation (in brackets) of the inference time (in second) of the original and converted DL models to ONNX and CoreML using different frameworks.
    }
    \label{tab:inference-time-summary}
   \begin{adjustbox}{width=0.9\textwidth,center}
    \begin{tabular}{l c | c|c|c||  c| c| c}

    \multirow{3}{*}{\textbf{Model}}& \multirow{3}{*}{\textbf{Framework}}&\multicolumn{3}{c}{\textbf{Loading Time}} &\multicolumn{3}{c}{\textbf{Inference Time}} \\ & & \multirow{2}{*}{\textbf{Original models}} & \multicolumn{2}{c}{\textbf{Converted models}}&\multirow{2}{*}{\textbf{Original models}} & \multicolumn{2}{c}{\textbf{Converted models}}\\
    & &  &\textbf{ONNX}&\textbf{CoreML}&&\textbf{ONNX}&\textbf{CoreML}\\ 
    
    \hline
     \multirow{2}{*}{\textbf{LeNet-5}}&\textbf{Keras}&0.21 (8.4E-03)&7.6E-04 (5.2E-04)&5.2E-02 (2.4E-02)&5.7E-02 (1.7E-02) & 7.6E-03 (2.5E-03) & 0.10(0.02)	\\
    
    &\textbf{PyTorch}&7.2E-04 (1.0E-02) &7.2E-04 (1.4E-03) &4.9E-02 (1.1E-02) &7.3E-03 (2.5E-03)	&6.9E-03 (2.7E-03)& 7.3E-02 (1.2E-02)	\\\hline

     \multirow{2}{*}{\textbf{ResNet-18}}&\textbf{Keras}&1.1 (3.4E-02) &8.2E-02 (5.0E-02) &3.0 (28.7) &1.8 (0.47)	& 1.5 (0.29) & 0.94 (0.12) \\
    
    &\textbf{PyTorch}&8.9E-02 (7.2E-03) &8.6E-02 (1.47E-02) &0.99 (0.15) &0.35 (0.06) & 0.15 (0.11)& 0.38 (0.09) \\\hline

     \multirow{2}{*}{\textbf{VGG}}&\textbf{Keras}&1.0 (0.83) &7.2E-02 (1.4E-02) &1.1 (0.23) &0.96 (0.18) & 7.6E-01 (1.4E-01)& 0.54 (0.05) \\
    
    &\textbf{PyTorch}&0.45 (0.03) &1.1 (0.21) &16 (4.2) & 371 (1020)&20.2 (5.2) &18.8 (5.2)	\\\hline
    
     \multirow{2}{*}{\textbf{LSTM}}&\textbf{Keras}&1.8 (0.58) &9.1E-02 (1.4E-02) &1.4 (0.1) &0.31 (0.36) & 0.32 (0.05) & 2.1 (0.14) 	\\
    
    &\textbf{PyTorch}&1.3E-02 (4.9E-03) &7.3E-03 (4.7E-03) &0.74 (0.39) & 3.7 (0.72) & 2.2 (0.62) & 2.3(0.68) \\\hline

     \multirow{2}{*}{\textbf{GRU}}&\textbf{Keras}&1.4 (0.15) &1.1E-01 (1.8E-02) &1.5 (3.4) &0.28 (0.4) & 0.32 (0.05) & 2.1 (0.16) \\
    
    &\textbf{PyTorch}&1.3E-02 (1.1E-03) &7.0E-03 (2.6E-03) &4.4 (1.1) &3.4 (1.0) & 2.2 (0.7)& 59 (139) \\
    
    \end{tabular}
    \end{adjustbox}

\end{table*}
\begin{table}[t]
    \centering
    \small
    \caption{The comparison of mean and standard deviation (in brackets) of the size (in MB) of original and converted DL models to ONNX and CoreML from the respective framework.
    }
    \label{tab:memory-summary}
   \begin{adjustbox}{width=\columnwidth,center}
    \begin{tabular}{l c |  c| c| c}

    \multirow{3}{*}{\textbf{Model}}& \multirow{3}{*}{\textbf{Framework}}&\multicolumn{3}{c}{\textbf{Model size}} \\ & & \multirow{2}{*}{\textbf{Original models}} & \multicolumn{2}{c}{\textbf{Converted models}}\\
    & & &\textbf{ONNX}&\textbf{CoreML}\\ 
    
    \hline
     \multirow{2}{*}{\textbf{LeNet-5}}&\textbf{Keras}&0.6019 (0) & 0.5727 (0)	&  0.5709(0)	\\
    
    &\textbf{PyTorch}&0.2523 (0)	&0.2481 (0)& 0.2476 (0)	\\\hline

     \multirow{2}{*}{\textbf{ResNet-18}}&\textbf{Keras}& 89.8 (0) & 44.7 (0) & 44.7 (0) \\
    
    &\textbf{PyTorch}&44.8 (0) & 44.7 (0) & 44.7 (0) \\\hline

     \multirow{2}{*}{\textbf{VGG}}&\textbf{Keras} & 71.9 (0) & 35.8 (1.07E-04)& 35.8 (6.24E-05)	\\
    
    &\textbf{PyTorch}&515.3 (0)&515.2 (0)&515.2 (0)	\\\hline
    
     \multirow{2}{*}{\textbf{LSTM}}&\textbf{Keras}&144.3 (0.2) &  48.1 (0.07)	& 48.2 (0.07) 	\\
    
    &\textbf{PyTorch}& 3.6837 (0)	&3.6833 (0)& 3.6767(0)		\\\hline

     \multirow{2}{*}{\textbf{GRU}}&\textbf{Keras}&240.0 (0.37)	&  47.94 (0.073) & 47.99 (0.074) 	\\
    
    &\textbf{PyTorch}&2.8276 (0)	& 2.8271 (0)& 133.9 (0)		\\
    
    \end{tabular}
    \end{adjustbox}

\end{table}

\begin{tcolorbox}[colback=blue!5,colframe=blue!40!black]
\textbf{Findings:} The difference of inference time among frameworks is significant. Moreover, \texttt{coremltools} is generally slower than ORT. Converted models are smaller in size, and this is intensified for RNNs.

\textbf{Challenges:} In few cases (specially for PyTorch), the inference process of converted models is very slow. This is an alert for DL software developers to be cautious on their deployed system that may perform poorly while switching from one framework to another. 
\end{tcolorbox}
\subsection{RQ3: Adversarial Robustness of Converted Models}
The adversarial robustness assessment was conducted in two parts: (1) assessing CNN models (image classification) and (2) assessing RNN models (NLP).

\textbf{CNN models:}
To assess the robustness of CNN models, we follow the methodology adopted by similar studies \cite{guo2019empirical}. We investigate the robustness of CNNs in terms of success rate against adversarial examples using two well-known adversarial attacks: Fast Gradient Sign Method (FGSM) \cite{goodfellow2014explaining} and Boundary Attack \cite{brendel2017decision}. To create adversarial examples, FGSM applies perturbation along the model's gradient. Boundary Attack, on the other hand, conducts a strong adversarial perturbation on the input first, then reduces the L2 norm distance of the perturbations while remaining adversarial. We start by picking 1000 images at random from MNIST and CIFAR-10 datasets that are correctly classified by the model both before and after conversion. These images will be used to attack the model prior to conversion using FGSM and Boundary Attack. These attack algorithms will generate new sets of synthetic images (i.e., adversarial samples per attack) referred to as $D_{syn-CNN}$. The latter will be used to compute (i) the success rate of attack algorithms on the model before conversion and (ii) the success rate of the converted model (through ONNX and CoreML). Since we have 10 instances for each model (trained and converted), the attack is performed on each instance and the average results are reported. We also compute misclassification of adversarial samples, i.e., the ratio of adversarial samples classified differently by original and converted models on $D_{syn-CNN}$. Furthermore, similar to Subsection \ref{accuracyConverted}, we measure the difference (i.e., the absolute error) between the last layers' outputs (i.e., predictions) of original and converted models over $D_{syn-CNN}$. All results are averaged over 10 instances per model.
In overall, we run 240 configurations of attacks for CNNs, i.e., 3 models $\times$ 2 frameworks $\times$ 2 types of conversions $\times$ 2 types of attacks $\times$ 10 times.

\begin{figure*}[t]
\center
\includegraphics[width=0.83\linewidth]{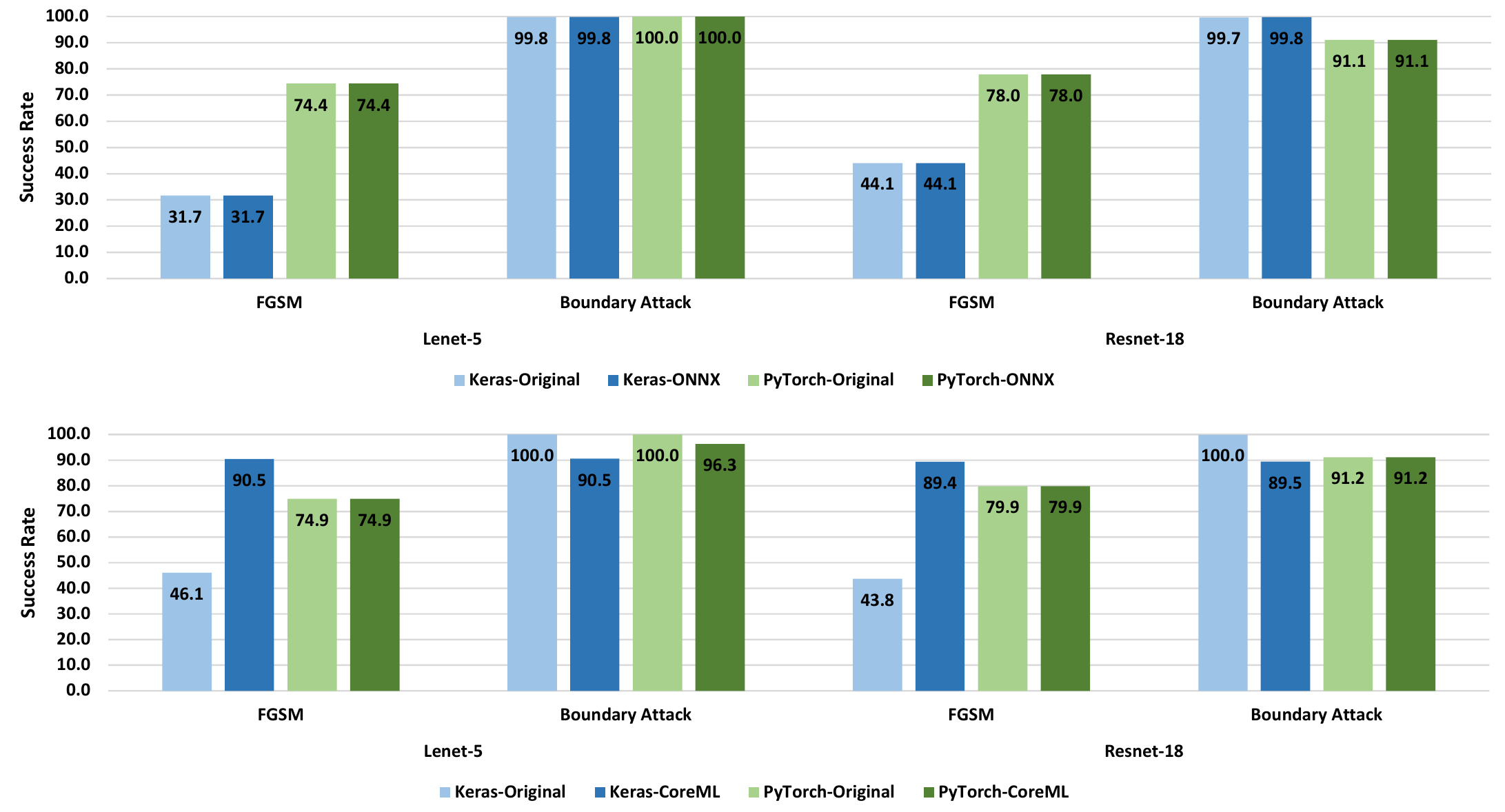}
\caption{The comparison of average success rate of different attacks on LeNet-5 and ResNet-18 models before and after conversion to ONNX (top) and CoreML (bottom).}
\label{fig:success_rate_CNN}
\end{figure*}

\begin{figure*}[t]
\center
     \begin{subfigure}[b]{0.45\textwidth}
         \centering
         \includegraphics[width=0.9\textwidth]{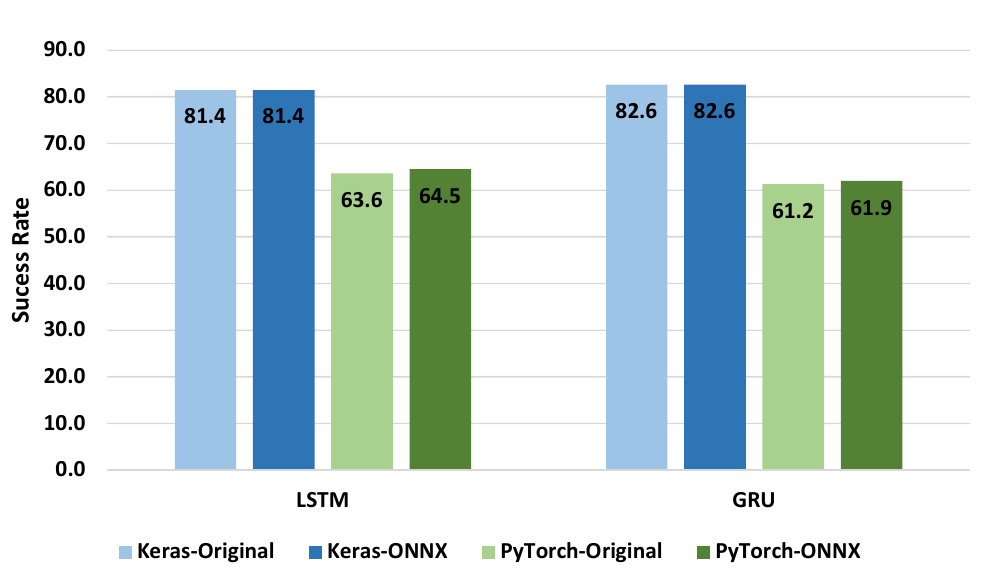}
         \caption{ONNX}
         \label{fig:RNN-ONNX}
     \end{subfigure}
     \hfill
     \begin{subfigure}[b]{0.45\textwidth}
         \centering
         \includegraphics[width=0.9\textwidth]{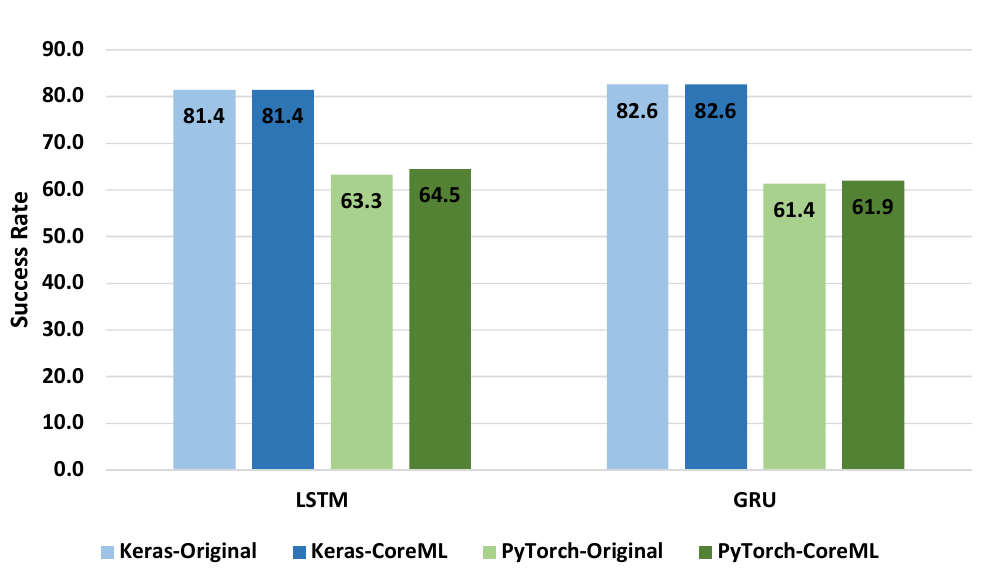}
         \caption{CoreML}
         \label{fig:RNN-CoreML}
     \end{subfigure}
     \hfill
\caption{The comparison of average success rate of different attacks on LSTM and GRU models before and after conversion to ONNX (left) and CoreML (right).}
\label{fig:success_rate_RNN}
\end{figure*}

\textbf{RNN models:}
Similar to CNNs, first we examine the robustness of our RNN models against adversarial examples in terms of success rate by leveraging the black-box population-based algorithm \cite{alzantot-2018}. This technique uses the combination of constrained word embedding distance and language model prediction score to narrow the search space. For the search algorithm, a well-known population-based metaheuristic algorithm, Genetic Algorithm, is adopted \cite{alzantot-2018}. We do not employ language model output in our experiments since the search space is manageable. We start by picking 1000 sentences at random from the IMDB that are correctly classified by the model both before and after conversion. Following \cite{alzantot-2018}, we limit the original input length to 10-100 words, removing out-of-vocabulary terms from the substitution set, and eliminating adversarial examples with modification rates of more than 25\%. We used this threshold since it was suggested originally for the black-box population-based algorithm by the authors \cite{alzantot-2018}. By having the set of adversarial samples, $D_{syn-RNN}$, similar to the CNN models, the success and misclassification rate over $D_{syn-RNN}$ between the original and converted models are reported. Moreover, the absolute error between the last layers' outputs (i.e., predictions) of original and converted models over adversarial samples ($D_{syn-RNN}$) are reported. All results are averaged over 10 instances per model. All results are averaged over 10 instances per model. In this experiment, we perform 80 configurations of attacks, i.e., 2 models $\times$ 2 frameworks $\times$ 2 types of conversions $\times$ 1 type of attack $\times$ 10 times. 

Figure \ref{fig:success_rate_CNN} illustrates the average success rate of different attacks against original and converted CNN models\footnote{We do not report results of the VGG model due to a problem in our experiments that made the results unreliable.}. The results are averaged over 10 instances per model. Boundary attack, compared to FSGM, achieves higher success rates on all CNN models, in some cases 100\%. Similar results were reported in other studies \cite{guo2019empirical} and the reason is that this attack is one of the most effective decision-based adversarial attacks. This indicates that both original and converted models are vulnerable against adversarial attacks. This is crucial for security of DL-based software, and therefore, defense mechanisms are desirable for trained DL models prior to deployment. For ONNX models, the success rates of both attacks remain almost the same before and after conversion. However, models trained with Keras are more vulnerable for both attacks as success rates are higher. There are two major observations in the case of CoreML:
\begin{enumerate}
    \item \textbf{PyTorch vs. Keras}: While PyTorch models demonstrate almost the same success rate after conversion, models trained by Keras behave differently after conversion: success rates are higher for converted models.
    \item \textbf{FGSM vs. Boundary}: Both LeNet-5 and ResNet-18, when attacked by FGSM, look are more vulnerable after conversion to CoreML, i.e., success rates jump from 46\% and 43\% to 90\% and 89\% respectively. In the case of Boundary attack, converted models reveal almost the same vulnerability level before and after conversion.
\end{enumerate}
\begin{table*}[t]
    \centering
    \large
    \caption{The comparison of absolute error and divergence (as percentage) of robustness evaluation of the converted CNN models from the respective framework to ONNX and CoreML.}
    \label{tab:robust-1}
   \begin{adjustbox}{width=0.75\linewidth}
    \begin{tabular}{l c c | c| c| c| c}
    
    \multirow{2}{*}{\textbf{Model}}& \multirow{2}{*}{\textbf{Attack}}&\multirow{2}{*}{\textbf{Framework}}&\multicolumn{2}{c|}{\textbf{Absolute Error}}&\multicolumn{2}{c}{\textbf{Misclassification (\%)}}\\
    & & &\textbf{ONNX}&\textbf{CoreML}  &\textbf{ONNX}&\textbf{CoreML}\\ 
    
    \hline
     \multirow{4}{*}{\textbf{LeNet5}}&\multirow{2}{*}{\textbf{FGSM}}&\textbf{Keras}& 2.0E-08 (2.0E-09) & 0.1712 (0.0011) & 0 (0) & 92.27 (1.9) \\
    
    & &\textbf{PyTorch} & 1.1E-06 (1.6E-07) & 0.0076 (0.0016) & 0 (0) & 0.02 (0.04)\\
    
    & \multirow{2}{*}{\textbf{Boundary}} &\textbf{Keras} & 1.2E-07 (6.9E-09) & 0.1512 (0.0016) & 0 (0) & 69.24 (7.67)\\
    
    & &\textbf{PyTorch}& 9.8E-07 (1.6E-07) & 0.0061 (0.0013) & 0 (0) & 1.83 (0.82) \\\hline
    
     \multirow{4}{*}{\textbf{ResNet-18}}&\multirow{2}{*}{\textbf{FGSM}}&\textbf{Keras} & 9.1E-07 (3.9E-08) & 0.1523 (0.001) & 0.1 (0.3162) & 89.01 (7.77)\\
    
    & &\textbf{PyTorch}& 1.9E-06 (1.6E-06) & 1.4E-06 (3.1E-08) & 0 (0) & 0 (0) \\
    
    & \multirow{2}{*}{\textbf{Boundary}} &\textbf{Keras} & 1.5E-06 (1.4E-08) & 0.1462 (0.0007) & 0.24 (0.18) & 87.25 (6.49) \\
    
    & &\textbf{PyTorch}& 3.7E-05 (7.6E-06) & 2.1E-05 (4.2E-06) & 0 (0) & 0 (0) \\
    
    
    
    
    
    
    \end{tabular}
    \end{adjustbox}

\end{table*}

\begin{table}[t]
    \centering
    \large
    \caption{Absolute error and divergence (as percentage) of robustness evaluation of the converted RNN models from the respective framework to ONNX and CoreML.}
    \label{tab:robust-2}
   \begin{adjustbox}{width=\linewidth}
    \begin{tabular}{l c | c | c| c| c}
    
    \multirow{2}{*}{\textbf{Model}}& \multirow{2}{*}{\textbf{Framework}}&\multicolumn{2}{c|}{\textbf{Absolute Error}}&\multicolumn{2}{c}{\textbf{Misclassification (\%)}}\\
    & &\textbf{ONNX}&\textbf{CoreML}  &\textbf{ONNX}&\textbf{CoreML}\\ 
    
    \hline
     \multirow{2}{*}{\textbf{LSTM}}&\textbf{Keras}& 1.1E-07 (2.0E-09) & 9.5E-08 (1.7E-08) & 0 (0) & 0 (0) \\
    
    &\textbf{PyTorch} & 0.0144 (0.0019) & 0.0144 (0.0017) & 2.75 (0.73) & 2.71 (0.63)\\
    
    \hline
    
     \multirow{2}{*}{\textbf{GRU}}&\textbf{Keras} & 1E-07 (2.2E-08) & 8.4E-08 (1E-08) & 0 (0) & 0 (0)\\
    
    &\textbf{PyTorch}& 0.0169 (0.0029) & 0.0168 (0.0027) & 36.3 (10.00) & 38.3 (9.96) \\
    
    \hline

    
    
    
    
    
    \end{tabular}
    \end{adjustbox}

\end{table}

The average success rate of different attacks against different RNN models (original and converted) are reported in Figure \ref{fig:success_rate_RNN}. The first observation is that ONNX (left) and CoreML (right) models behave similarly and no significant contrast is noticed for both LSTM and GRU. In general, the vulnerability level (in terms of attack’s success rate) of converted models remains almost the same after conversion and no major difference is observed. Interestingly, using the same runtime configurations, PyTorch models exhibit more robust behavior compared to Keras models. This phenomenon is intensified for GRU, where the success rates of the attacks drop from 82\% (Keras) to near 20\% (PyTorch).

Table \ref{tab:robust-1} and \ref{tab:robust-2} show the results of robustness evaluation of different models per attack by reporting the models' prediction error and misclassification rate over adversarial examples for CNN and RNN models respectively. The absolute errors and misclassification are relatively low for ONNX models in general (actually, zero classification divergence in many cases). So one may conclude that the adversarial robustness of trained models remains almost the same after conversion to ONNX. However, CoreML models behave distinctly as the absolute errors are considerable for all frameworks and attacks implying that converted models to CoreML behave differently compared to original models under adversarial attacks. For example, LeNet-5 trained in Keras under FGSM, displays the misclassification rate of 92\%, meaning that almost all adversarial samples were classified differently after conversion. The only exception is FGSM attack on PyTorch models, where both ONNX and CoreML models do not reveal any significant difference compared to original ones. For CNNs, no major difference is observed when comparing frameworks (Keras vs. PyTorch) or types of attacks (FGSM vs. boundary attack). However, in the case of RNNs, the error and miscalssification rate of models trained using PyTorch are higher than those trained with Keras, but the models are still at the similar level of robustness after conversion. According to our results, deployment of CoreML models results in poor software quality since the system makes different decisions for the same adversarial input. This must be taken into account by the development and deployment teams while more investigation is necessary. Also, future works can propose efficient fault detection and verification techniques to support debugging or testing of the DL models during the conversion and the deployment process. 

To have a global understanding of the ability of each conversion format in preserving the adversarial robustness of models, we also present results of a robustness indicator metric introduced in \cite{guo2019empirical}. Formally, we define the following equations to quantify the robustness variation of a converted model under attacks:
\begin{equation}
\centering
\begin{split}
\label{eqn:eq4}
    R(m_i, c_j) = P(m_i , c_j, A_1) + . . . + P(m_i , c_j, A_k), k \geq 1, \\ \text{with}\,\, P(m_i, c_j, A) = 
\begin{cases}
\frac{\left | S_{c_j}^{m_i,A} - S^{m_i,A} \right | - min}{max - min}\;\text{ if }  max > min\\ 
0 \text{ if } \;max = min
\end{cases}
\end{split}
\end{equation}
where $m_1$, ..., $m_n$ represent the $n$ models trained from frameworks under evaluation, and $A_k$ represents the k types of attacks. $S^{m_i,A}$ reflects the average success rate of attack $A$ on model $m_i$ before conversion. $S_{c_j}^{m_i,A}$ represents the average success rate of attack $A$ on model $m_i$ after conversion using $c_j$. Finally, $min$ and $max$ indicate the minimum and maximum of the difference in success rate before and after conversion of all models under attack $A$, respectively.

We compute the final robustness indicator $R(m_i, c_j)$ with Equation \ref{eqn:eq4}, which quantifies the robustness variation after the conversion $c_j$ of model $m_i$ in terms of $k$ attacks $A1, . . . , A_k$. The smaller the value $R(m_i, c_j)$ is, the better the conversion format $c_j$ conserves robustness. In this study, $m_1$, and $m_2$ represent LeNet-5 and ResNet-18 models trained by Keras, and PyTorch respectively. $A_1$ and $A_2$ indicate FGSM attack and Boundary attack, respectively. Finally, $c_1$ and $c_2$ represent ONNX and CoreML as well.

\begin{table}[t]
    \centering
    \large
    \caption{Robustness indicator of CNN models after being converted using ONNX and CoreML.}
    \label{tab:robust-3}
   \begin{adjustbox}{width=0.85\linewidth}
    \begin{tabular}{c | c | c| c| c}
    
    \multirow{2}{*}{\textbf{Framework/Model}}
    &\multicolumn{2}{c|}{\textbf{ONNX}}&\multicolumn{2}{c}{\textbf{CoreML}}\\
    &\textbf{LeNet5}&\textbf{ResNet-18}  &\textbf{LeNet5}&\textbf{ResNet-18}\\ 
    
    \hline
     \textbf{Keras}& 0 & 0.12 & 1.87 & 2 \\
     
     \hline
    
    \textbf{PyTorch} & 0 & 0 & 0.35 & 0\\

    \hline
    
    \end{tabular}
    \end{adjustbox}

\end{table}

Using Equation \ref{eqn:eq4}, we discover that ONNX conserves the adversarial robustness of models better than CoreML. This is shown in Table \ref{tab:robust-3}, where the robustness indicator is zero for all ONNX models except ResNet-18 with Keras (0.12). CoreML, on the other hand, performed poorly in terms of preserving CNN models' adversarial robustness. The robustness indicator returns high values for CoreML settings, such as 1.87 and 2 for LeNet5 and ResNet-18 trained by Keras. In addition, Table \ref{tab:robust-3} reveals that models trained by PyTorch have higher adversarial robustness than models trained in Keras. In reality, the average robustness indicator score of all Keras models (using both conversion formats) is 0.99, whereas the average PyTorch model robustness indicator score is 0.09.  


\begin{tcolorbox}[colback=blue!5,colframe=blue!40!black]
\textbf{Findings:} ONNX models generally are almost at the same level of adversarial robustness compared to original models. However, models trained with PyTorch are showing more robustness. Some CoreML models look very vulnerable after conversion as well.

\textbf{Challenges:} Since both ONNX and CoreML are vulnerable against adversarial attacks, robust deployment of a converted model is challenging.
\end{tcolorbox}

\subsection{Threats to Validity}
The selected DL models/datasets might not be complete and representative of state-of-the-art practices in the DL community. So, our findings are not general for all situations. However, we select models with CNN and RNN architectures from various domains, ranging from image classification to textual sentiment analysis. Moreover, since many of the operators in DL models are common and already exist in the well-known models studied by our paper, we do not think studying big models will impact our findings much (as they share lots of common operators). 
The datasets, therefore, contain diverse data, including gray, color images and textual review, to reduce such a threat. Although we select two popular DL frameworks in our study, the results can be extendable to other frameworks. However, the focus of this paper is not on the multi-version evolution, but on revealing challenges/issues that developers and researchers need to consider in development and deployment processes of DL-based software.

\section{Discussion} \label{discussion}
The conversion of DL models are useful for simplifying the maintenance and evolution of DL systems since they allow models to be trained in the preferred framework and run elsewhere, on the cloud/edge as needed. In fact, optimizing the DL model for inference can be difficult since one needs to tune the model and corresponding library to optimize hardware capabilities. This will be very complicated and expensive when one has multiple models from a variety of frameworks to be deployed on different platforms (e.g., cloud/edge, CPU/GPU) due to different capabilities and characteristics of frameworks/platforms.

From the results presented, we can see that Keras models generally behave quite differently in terms of the prediction accuracy, the inference time and memory usage when converted to ONNX and CoreML formats. The converted models (i.e., ONNX or CoreML) in their runtime environments have faster inference time and take less memory space compared to the original. Keras models tend to take quite a larger memory space compared to the similar model in PyTorch. The ONNX Runtime is claimed to be designed to considerably increase performance over multiple models \cite{ONNX:runtime}. Also, it is not surprising to confirm that models in CoreML runtime tools perform quite poorly (higher prediction error compared to ONNX), slower (higher inference time), take much less memory space (small model size), and are more vulnerable to attacks (especially for the models trained in Keras). This is expected because CoreML is explicitly optimized for on-device performance, as explained by its developers. Based on our findings presented in this paper, we appeal to the DL developers to be cautious on deployed models that may 1) perform poorly while switching from one framework to another, 2) have challenges in robust deployment, or 3) run slowly, leading to poor quality of deployed DL-based software that can lead to lower user satisfaction. This also includes DL-based software maintenance tasks, such as bug prediction.

\section{Conclusion} \label{conclusion}
In this paper, we present the first empirical study on converting trained DL models into deployable formats. Five DL models, selected from CNNs and RNNs, were trained on three well-known datasets in the DL community. Then, we converted models into ONNX and CoreML. We evaluated prediction accuracy, size, inference time, and adversarial robustness of converted models and compared to original ones. Our results reveal differences in original and converted models. 
From the results reported, it could be interesting to comprehensively explore the impact of the different hyperparameters on the converted model. However, this may require more experiments to have a concrete conclusion. We need to expand our analysis by exploring the impact of different input sizes and shapes 
when converting from the DL frameworks to ONNX or CoreML. It is still interesting to tune our analysis to compare the results across the selected runtime environments for converted models 
and also report the issues across these environments.




\section*{Acknowledgment}
We acknowledge the support from the following organizations and companies: Fonds de Recherche du Québec (FRQ), Natural Sciences and Engineering Research Council of Canada (NSERC), Canadian Institute for Advanced Research (CIFAR) and Huawei Canada. However, the findings and opinions expressed in this paper are those of the authors and do not necessarily represent or reflect those organizations/companies.

\balance
\bibliography{sample-base}
\bibliographystyle{ieeetr}

\end{document}